\PassOptionsToPackage{unicode}{hyperref}
\PassOptionsToPackage{hyphens}{url}
\PassOptionsToPackage{dvipsnames,svgnames,x11names}{xcolor}
\documentclass[12pt]{article}

\usepackage{amsmath,amssymb}
\usepackage{iftex}
\ifPDFTeX
  \usepackage[T1]{fontenc}
  \usepackage[utf8]{inputenc}
  \usepackage{textcomp} 
\else 
  \usepackage{unicode-math}
  \defaultfontfeatures{Scale=MatchLowercase}
  \defaultfontfeatures[\rmfamily]{Ligatures=TeX,Scale=1}
\fi
\usepackage{lmodern}
\ifPDFTeX\else  
\fi
\IfFileExists{upquote.sty}{\usepackage{upquote}}{}
\IfFileExists{microtype.sty}{
  \usepackage[]{microtype}
  \UseMicrotypeSet[protrusion]{basicmath} 
}{}
\makeatletter
\@ifundefined{KOMAClassName}{
  \IfFileExists{parskip.sty}{%
    \usepackage{parskip}
  }{
    \setlength{\parindent}{0pt}
    \setlength{\parskip}{6pt plus 2pt minus 1pt}}
}{
  \KOMAoptions{parskip=half}}
\makeatother
\usepackage{xcolor}
\makeatletter
\ifx\paragraph\undefined\else
  \let\oldparagraph\paragraph
  \renewcommand{\paragraph}{
    \@ifstar
      \xxxParagraphStar
      \xxxParagraphNoStar
  }
  \newcommand{\xxxParagraphStar}[1]{\oldparagraph*{#1}\mbox{}}
  \newcommand{\xxxParagraphNoStar}[1]{\oldparagraph{#1}\mbox{}}
\fi
\ifx\subparagraph\undefined\else
  \let\oldsubparagraph\subparagraph
  \renewcommand{\subparagraph}{
    \@ifstar
      \xxxSubParagraphStar
      \xxxSubParagraphNoStar
  }
  \newcommand{\xxxSubParagraphStar}[1]{\oldsubparagraph*{#1}\mbox{}}
  \newcommand{\xxxSubParagraphNoStar}[1]{\oldsubparagraph{#1}\mbox{}}
\fi
\makeatother

\usepackage{longtable,booktabs,array}
\usepackage{calc} 
\usepackage{etoolbox}
\makeatletter
\patchcmd\longtable{\par}{\if@noskipsec\mbox{}\fi\par}{}{}
\makeatother
\IfFileExists{footnotehyper.sty}{\usepackage{footnotehyper}}{\usepackage{footnote}}
\makesavenoteenv{longtable}
\usepackage{graphicx}
\makeatletter
\def\maxwidth{\ifdim\Gin@nat@width>\linewidth\linewidth\else\Gin@nat@width\fi}
\def\maxheight{\ifdim\Gin@nat@height>\textheight\textheight\else\Gin@nat@height\fi}
\makeatother
\setkeys{Gin}{width=\maxwidth,height=\maxheight,keepaspectratio}
\makeatletter
\def\fps@figure{htbp}
\makeatother

\makeatletter
\@ifpackageloaded{caption}{}{\usepackage{caption}}
\AtBeginDocument{%
\ifdefined\contentsname
  \renewcommand*\contentsname{Table of contents}
\else
  \newcommand\contentsname{Table of contents}
\fi
\ifdefined\listfigurename
  \renewcommand*\listfigurename{List of Figures}
\else
  \newcommand\listfigurename{List of Figures}
\fi
\ifdefined\listtablename
  \renewcommand*\listtablename{List of Tables}
\else
  \newcommand\listtablename{List of Tables}
\fi
\ifdefined\figurename
  \renewcommand*\figurename{Figure}
\else
  \newcommand\figurename{Figure}
\fi
\ifdefined\tablename
  \renewcommand*\tablename{Table}
\else
  \newcommand\tablename{Table}
\fi
}
\@ifpackageloaded{float}{}{\usepackage{float}}
\floatstyle{ruled}
\@ifundefined{c@chapter}{\newfloat{codelisting}{h}{lop}}{\newfloat{codelisting}{h}{lop}[chapter]}
\floatname{codelisting}{Listing}

\makeatother
\makeatletter
\@ifpackageloaded{caption}{}{\usepackage{caption}}
\@ifpackageloaded{subcaption}{}{\usepackage{subcaption}}
\makeatother

\ifLuaTeX
  \usepackage{selnolig}  
\fi
\usepackage[]{natbib}
\usepackage{bookmark}

\IfFileExists{xurl.sty}{\usepackage{xurl}}{} 
\hypersetup{
  pdftitle={Title},
  pdfauthor={Author 1; Author 2},
  pdfkeywords={3 to 6 keywords, that do not appear in the title},
  colorlinks=true,
  linkcolor={blue},
  filecolor={Maroon},
  citecolor={Blue},
  urlcolor={Blue},
  pdfcreator={LaTeX via pandoc}}

\newcommand{\anon}{1}

\newcommand{\diag}{\mathrm{diag}}

\begin{document}

\def\spacingset#1{\renewcommand{\baselinestretch}%
{#1}\small\normalsize} \spacingset{1}


\if1\anon
{
  \title{\bf The Morgan-Pitman Test of Equality of Variances and its Application to Machine Learning Model Evaluation and Selection}
  \author{Argimiro Arratia$^{(1)}$, Alejandra Caba\~na$^{(2)}$, \\ Ernesto Mordecki$^{(3)}$, Gerard Rovira-Parra$^{(1)}$\thanks{
    A. Caba\~na was supported by  grant PID2021-123733NB-100, Ministerio de Ciencia, Innovaci\'on y Universidades, Spain.
\newline
G. Rovira-Parra was supported by grant J03128 - PID2022-143299OB-I00, Ministerio de Ciencia e Innovaci\'on, Spain.}\\ \\
       $^{(1)}$Department of Computer Science\\
       Universitat Polit\`ecnica de Catalunya\\
       Barcelona 08034, SPAIN. \{argimiro.arratia, gerard.rovira.parra\}@upc.edu\\
       $^{(2)}$Department of Mathematics\\
       Universitat Aut\`onoma de Barcelona\\
       Bellaterra  08193, SPAIN. AnaAlejandra.Cabana@uab.cat\\
 	 $^{(3)}$Centro de Matem\'atica, Facultad de Ciencias\\
	Montevideo 11400, URUGUAY. mordecki@cmat.edu.uy\\
    }
     \date{}
  \maketitle
} \fi

\if0\anon
{
  \bigskip
  \bigskip
  \bigskip
  \begin{center}
    {\LARGE\bf The Morgan-Pitman Test of Equality of Variances and its Application to Machine Learning Model Evaluation and Selection}
\end{center}
  \medskip
} \fi

\bigskip
\begin{abstract}
Model selection in non-linear models often prioritizes performance metrics over statistical tests, limiting the ability to account for sampling variability. We propose the use of a statistical test to assess the equality of variances in forecasting errors. The test builds upon the classic Morgan-Pitman approach, incorporating enhancements to ensure robustness against data with heavy-tailed distributions or outliers with high variance,
plus a strategy to make residuals from machine learning models
statistically independent.
Through a series of simulations and real-world data applications, we demonstrate the test's effectiveness and practical utility, offering a reliable tool for model evaluation and selection in diverse contexts.
\end{abstract}

\noindent%

\spacingset{1.8} 

\section{Introduction}

The explosive development of artificial intelligence,  
supported by the emergence of a large number of new methods, raises a wide range of statistical questions including the validation, interpretability, and quality estimation of models.
Two primary and interrelated procedures dominate the tasks of model selection and evaluation in today's landscape. The first is the division of the data into training and testing subsets, which is typically done by random partitioning. The second is cross-validation, a more systematic approach through leave-one-out validation or $k$-fold splitting~\citep{stone}. Both methods aim to estimate model performance while reducing overfitting and bias introduced by specific data partitions. 

In this sense, statistical tools that are both straightforward and effective for assessing model quality and guide model selection are needed. These tools must ensure that practitioners who analyze the same data with the same model can achieve consistent and reproducible results.
In line with this methodological principle, we propose a simple test to validate extended machine learning practices. The proposed test evaluates whether the variances of the forecast errors between different models are statistically equivalent.

Unlike traditional metrics that focus exclusively on predictive accuracy, this test offers a complementary perspective by assessing the stability and reliability of model predictions through variance comparison. The test is built upon the classic test of equality of variances introduced by Morgan~\citep{morgan} and independently by Pitman~\citep{pitman}, improved to make it robust against sampling from data with heavy-tailed distributions, observations with large variance or small data sets.

The primary intended application of this statistical test is the comparison of nested models in line with the idea of the $F$-test for linear models, where the original model is more complex but shares the same structure as the simpler alternative. If the test does not reject the equality of variances, there is no significant loss in adopting the simpler model. This suggests that the additional complexity of the original model does not offer any significant improvement, potentially reflecting overfitting. The nested model, in contrast, captures the same relevant information while offering greater generalizability. Hence, by the principle of parsimony, the simpler model is preferred.
 On the other hand, if the test rejects equality of variances, we can assume strict inequality due to the nesting property.
 In this regard, a context where it could be applied is in the reduction of input features in a non-linear model.

However, in the realm of data-driven statistics, as is the case of machine learning modeling, 
out of the class of linear models with i.i.d. Gaussian errors, it is not likely to obtain independent residuals, a necessary condition for the basic Morgan-Pitman equality of variances test. A $k$-fold cross-validation approach will not produce independent residuals, as noted by \citep{bengio2003}. Thus, as a further enhancement of this test, to make it applicable in machine learning model selection, we propose a strategy to make residuals statistically independent. This is explained in Section \ref{NestedModels}.

A test for equality of variances can find general applications in the practice of model
selection and evaluation
when comparing different models, not necessarily from the same family,
and further offering insights that standard accuracy-based metrics may overlook.

The mean square error of a prediction $\widehat{Y}$ of a target $Y$ and the variance of the error $\epsilon = Y - \widehat{Y}$ are related by: 
$$MSE = \mathbb{E}[(Y - \widehat{Y})^2] = Var(\epsilon) + (\mathbb{E}[\epsilon])^2.$$
If the prediction is unbiased (i.e. $\mathbb{E}[\epsilon] = 0$), then $MSE = Var(\epsilon)$.
Therefore, among models exhibiting statistically equivalent variance assuming they are unbiased or any bias has been removed (for instance, through bootstrapping) it is reasonable, in terms of generalization and genuine learning of the data-generating function (as opposed to merely overfitting the observed data), to favor the simpler model, which may not necessarily correspond to the one with the lowest mean squared error. Hence, in the general practice of model selection, we propose to test simultaneously for equality of variance of residuals and unbiasedness, or at least equality of biases.

The paper is organized as follows. In Section \ref{RelatedLiterature}, we review the relevant literature, provide additional motivation for the test, and trace its historical development. 
In Section \ref{ElTest}, we define the test of equality of variances. 
Section \ref{NestedModels} introduces the class of nested models on which we will focus, while Section \ref{experiments} details a series of experiments that apply the test. Finally, we summarize some conclusions in Section \ref{sec:conclusion}.

\section{Background}\label{RelatedLiterature}    
When evaluating a model's predictive performance, the typical approach relies on point estimates of forecast errors, often measured using loss functions such as mean absolute error, mean squared error, or root mean squared error. 
In general machine learning practice, the model that achieves the lowest out-of-sample error is commonly selected as the best performing~\citep{StatLearnBook}. 
However, these measures capture only the average precision and do not provide information on the statistical significance of observed differences. In other words, accuracy metrics do not represent formal statistical tests as they fail to determine whether a model's superior performance is statistically significant or merely a consequence of sampling variability.

Cross-validation is another popular methodology for evaluating and comparing models in machine learning (cf. \citep[Ch. 5]{StatLearnBook}).
Although cross-validation can help evaluate the stability of estimators across subsets of the data, it does so disregarding statistical significance; hence, it is harder to discern whether the observed effect is genuinely representative of an underlying pattern or merely an artifact of random chance. Statistical significance tests inform decision makers on whether the observed results are likely to have an impact or are worth acting on. Without such considerations, decisions based solely on point estimates can lead to suboptimal outcomes, particularly in applications where interventions involve substantial costs or risks.

Formal statistical tests have been proposed to compare the performance of two learning algorithms, in order to provide more reliable estimates of the variation attributable to the choice of training set than those produced by $k$-fold cross-validation-based tests. An example of such a test is the $5\times 2$-cv paired $t$-test \cite{dietterich1998}. However, despite employing five replications with two non-overlapping training sets, this test suffers from the issue that the error rate differences it relies on are not fully independent. This is because these differences are measured on opposite folds of a 2-fold cross-validation. Moreover, this test compares algorithms based solely on accuracy-like metrics such as error rate or mean error, failing to capture other important aspects such as robustness. A further significant drawback is that the algorithms under comparison are trained on sets that are only half the size of the available data.

In the context of neural networks, model selection generally involves determining the optimal architecture and identifying the most relevant input variables. Feature selection for neural networks often relies on computing partial derivatives of a relevance measure with respect to input variables, followed by multiple tests to assess significance~\citep{Horel20,RP15}.
This method offers benefits, but presents challenges regarding interpretability, computational complexity, and the risk of overfitting from multiple tests. For example, interpreting these relevance measures in deep neural networks is complex, as interlayer interactions complicate the attribution of importance to specific features. Additionally, the computational cost of relevance-based methods increases substantially for large neural networks or datasets, which may limit their applicability in real-world scenarios.

In the field of econometric modelling, some formal tests of predictive accuracy have been developed, in particular for time series data, where observations are typically dependent. The Diebold-Mariano (DM) test~\citep{DM95} is one of the most widely adopted tests for significant differences in forecast accuracy by examining paired residuals over time. Although the DM test was initially designed to compare predictive accuracy between two sets of forecasts, its use has been abused as a `fully-articulated' model comparison tool, where other simpler and more powerful standard full-sample comparison tools exist (see the profound critique by one of the authors in \citep{Die15}).

Although accuracy is undoubtedly a central metric, variance in forecast errors is an equally critical measure for model evaluation. Models with lower error variance are more consistent in their predictions, making them potentially more reliable in practical settings. A formal test for residual variance allows practitioners to determine whether observed differences in stability are statistically significant, which could be a valuable criterion in cases where model stability is prioritized. 

The proposed test is applicable to a wide variety of data types, extending beyond the time series emphasis of the DM test. It serves as a complementary tool for the machine learning community, enabling a more comprehensive approach to model evaluation that accounts for model stability based on their predictions, without relying solely on point measures that may arise from random variation. By performing formal statistical tests to detect significant differences in residual variances, researchers can make more informed decisions about the robustness and consistency of models. This approach supports the adoption of a more generalized model, thus improving the reliability of predictive models across diverse applications.

\section{An Improved Morgan-Pitman Test with Hete\-ros\-ke\-dastic Consistent Methods}\label{ElTest}
Consider an arbitrary random vector $(X,Y)$ with finite covariance matrix,
$$
\left[
\begin{array}{cc}
\sigma_X^2 & \sigma_{XY}\\
\sigma_{XY} & \sigma_Y^2
\end{array}
\right],
$$
and a sample of i.i.d. vectors $(X_1,Y_1),\dots,(X_n,Y_n)$ distributed as $(X,Y)$.
A well-known technique due to Morgan \citep{morgan} and Pitman \citep{pitman} to test the equality of variance
of $X$ and $Y$ is based on the fact that
their variances are equal if and only if
the covariance of $U = X+Y$ and $V=X-Y$ is zero. 
Thus, they propose substituting the null hypothesis of equality of variances of $X$ and $Y$,
\begin{equation}\label{vareq}
H_0 : \sigma_X^2 = \sigma_Y^2,
\end{equation}
by the equivalent null hypothesis of zero correlation of $U$ and $V$,
\begin{equation}\label{rhoUV}
H_0 : \rho_{UV} = 0.
\end{equation}
Here, $\rho_{UV}$ is the Pearson correlation between $U$ and $V$, estimated by its empirical counterpart:
$$
\hat{\rho}_{UV}=\frac{\displaystyle\sum_{i=1}^{n}(U_i - \bar{U})(V_i - \bar{V})}{
\left(\vphantom{\sum^n}\displaystyle\sum_{i=1}^{n}(U_i - \bar{U})^2\right)^{1/2}
\left(\vphantom{\sum^n}\displaystyle\sum_{i=1}^{n}(V_i - \bar{V})^2\right)^{1/2}},
$$
see \cite[(9.7.1)]{hogg}.
The test statistic is 
\begin{equation}\label{t-test}
T_{UV} = \hat{\rho}_{UV}\sqrt{\frac{n-2}{1-(\hat{\rho}_{UV})^2}},
\end{equation}
where $n$ is the sample size. 
This statistic is distributed as a Student's $t$-distribution with $n-2$ degrees of freedom
when the data are normally distributed, and even when only one of the
variables is normal and they are independent (see \cite[Remark 9.7.1]{hogg}).
Observe that as $\rho_{UV}=\frac{\sigma_X^2 - \sigma_Y^2}{\sigma_U \sigma_V}$, 
the null \eqref{vareq} is equivalent to \eqref{rhoUV},
and we can also test $\sigma_X^2 \lessgtr \sigma_Y^2$ by testing $\rho_{UV}\lessgtr 0$.

However, this straightforward method performs poorly when sampling from heavy-tailed distributions
(even with finite second moments), 
as reported by Wilcox and supported by simulation studies \citep{wilcox}.
Another issue to be taken into account is heteroscedasticity,
i.e. the variance of the errors varying across observations. 
Long and Ervin in \citep{LE00} suggest, based on simulation studies,
that methods that correct for heteroskedasticity are essential for prudent data analysis, 
and strongly recommend using tests based on heteroskedasticity-consistent covariance matrix (HCCM) estimators. 
The first and most commonly used form of the HCCM estimator is known as HC0, proposed by White in 1980~\citep{white80}, and subsequent modified versions
were proposed by other researchers to improve the behavior of HC0 on small samples and adjust for the possible influence of observations with large variance (an account of the development and formal definitions of these HCCM estimators is given in \citep{LE00}).
In this regard, we follow the guidance in \cite{wilcox}, which indicates that Cribari-Neto's HC4 correction \citep{HC4} performs well under heteroskedasticity, large variance, and small sample conditions.

According to the previous considerations, given two competing predictive models, say model $A$ and model $B$, acting on the same dataset, the statistical procedure to assess the significance of equality of the variances of their predictive errors is the following:
\begin{enumerate}
\item Calculate the out-of-sample prediction errors for model $A$, given by $\{\epsilon^A_i\}$, as well as those of model $B$, given by $\{\epsilon^B_i\}$, $i = 1, \ldots, n$.
\item For $i = 1, \ldots, n$: 
\begin{quote}
Define $U = \{U_i\}$, with $U_i = \epsilon^A_i + \epsilon^B_i$.\\
Define $V = \{V_i\}$, with $V_i = \epsilon^A_i - \epsilon^B_i$.
\end{quote}

\item Define the design matrix $M = \left(\begin{array}{cc}1 & U_1 \\\vdots & \vdots \\1 & U_n\end{array}\right)$ of the general regression model of $V$ on $U$, that is, $V = M\beta + \epsilon$, where $\epsilon$ is a vector of random errors. 
\item Compute $C = (M^tM)^{-1}$, where $M^t$ stands for transpose of $M$.
\item Compute $\hat{\beta} = CM^tV$, the Ordinary Least Squares (OLS) estimator of $\beta$.
\item Compute $\hat{\epsilon}=(\hat{\epsilon}_1\ldots, \hat{\epsilon}_n) = V-M\hat{\beta}$ the $n$-vector of residuals.
\item Compute the correction for the covariance matrix of $\epsilon$:
\begin{eqnarray*}
    H &=& (h_1, \ldots, h_n) = \diag(MCM^t)^{-1}\\
    \bar{h}  &=& \frac{1}{n}\sum_{i=1}^{n} h_{i} \\
    d_{i} &=& \min(4,\frac{h_{i}}{\bar{h}}), i = 1, \ldots, n\\
    \Omega &=& \diag(\hat{\epsilon}_1^2(1-h_1)^{-d_1}, \ldots,\hat{\epsilon}_n^2(1-h_n)^{-d_n} )
\end{eqnarray*}
\end{enumerate}

Then Cribari-Neto's HC4 estimator for this setup is
$${\cal S} = CM^t\Omega MC$$ 
Observe that $\mathcal{S}$ is a $2 \times 2$ matrix.
The element ${\cal S}_{2,2}$ corresponds to the estimated squared standard error of $\hat{\beta_1}$.

Since testing for (\ref{rhoUV}) is equivalent to testing for $\beta_1=0$, we compute the $p$-value  as the probability that a $t$-distribution with $n-2$ degrees of freedom is greater than $\frac{\| \hat{\beta_1}\|}{\sqrt{{\cal S}_{2,2}}}$ for a two-sided test. One-sided tests can be considered as well.

\section{Nested Models}\label{NestedModels}
When comparing different models, 
it is necessary not only to take into account the variance of the residuals
but also the complexity of the model. This is done in different ways depending on the
strategy, some classical references being
the Akaike information criterion (AIC) proposed in the framework of time series \citep{akaike},
the Bayesian information criterion (BIC) \citep{bic},
or the LASSO criterion (Least Absolute Shrinkage and Selection Operator)
proposed in Geophysics \citep{lassog}, and also by Tibshirani \citep{lasso}.

To take this fact into account, we consider \textit{nested models}
in such a way that the nested model has the same form of the original model but with only a subset of its parameters. 
Consider a model $f$ such that
$$
Y=f(X;\alpha)+\epsilon_{\alpha}.
$$
Here, the parameters of the model are
$\alpha=(\alpha_1,\dots,\alpha_n)$ 
and we assume that $X$ and $\epsilon_{\alpha}$ are independent.
A nested model complies with
$$
Y=f(X;\alpha')+\epsilon_{\alpha'},
$$
for $\alpha' \subseteq \alpha$ and we also assume that $X$ and $\epsilon_{\alpha'}$ are independent.

If the models are estimated by minimizing the least squares criterion, in the sense that
$$
\widehat\alpha=\arg\min_{\alpha} \mathbb{E}\left[(
Y-f(X;\alpha))^2
\right]
$$
and
$$
\widehat {\alpha'}=\arg\min_{\alpha'} \mathbb{E}\left[(
Y-f(X;\alpha'))^2
\right],
$$
then, since the search space for the infimum in the first model contains  the search space for the nested model, 
$$
\mathbb{E}\big[(Y-f(X;\widehat\alpha))^2\big]\leq \mathbb{E}\big[(Y-f(X;\widehat{\alpha'}))^2\big].
$$
Equivalently, provided the residuals are centered, this amounts to:
\begin{eqnarray}\label{eq_var}
    \sigma^2_{\epsilon_{\widehat\alpha}}\leq \sigma^2_{\epsilon_{\widehat{\alpha'}}}.
\end{eqnarray}
Note that the independence of the residuals of each model is not guaranteed, and this is a necessary condition for the applicability of the test. To obtain such residuals, a $k$-fold cross-validation approach is not a valid strategy, as noted by \citep{bengio2003}. Therefore, in addressing the within-block ($\omega$) and between-blocks ($\gamma$) covariances described therein, bootstrap samples can be drawn from the original dataset, with a single out-of-bag sample used as the test set to eliminate shared dependencies among residuals, which arise from being generated by the same trained model. This out-of-bag sample shall be randomly selected taking into account a record of previously selected samples to ensure that the selection is as evenly distributed among the samples as possible.

\subsection{Feature Selection as a Nested Model Problem}

There are many situations in which the search for a model
uses nested models. As an example, consider a feedforward neural network with \( L \) hidden layers.
Given an input vector \( \mathbf{x} \in \mathbb{R}^d \), 
we represent the neural network as a composition of functions.
To do this, define:
\begin{itemize}
\item
 \( \mathbf{W}^{[\ell]} \in \mathbb{R}^{h_\ell \times h_{\ell-1}} \) is the weight matrix for the hidden layer \( \ell \), where \( h_{\ell-1} \) is the number of neurons in the previous layer (for the input layer, \( h_0 = d \)).
\item \( \mathbf{b}^{[\ell]} \in \mathbb{R}^{h_\ell} \) is the bias vector for layer \( \ell \).
\item \( \mathbf{a}^{[\ell]} \in \mathbb{R}^{h_\ell} \): the activation vector for layer \( \ell \), with \( \mathbf{a}^{[0]} = \mathbf{x} \).
\end{itemize}
The output of each layer for $\ell=1,\dots,L$ is calculated as
\[
\mathbf{z}^{[\ell]} = \mathbf{W}^{[\ell]} \mathbf{a}^{[\ell-1]} + \mathbf{b}^{[\ell]},
\qquad
\mathbf{a}^{[\ell]} = g^{[\ell]}(\mathbf{z}^{[\ell]})
\]
and
$$ 
f^{[L+1]}(\mathbf{a}^{[L]}) = g_{\text{out}}(\mathbf{W}^{[L+1]} \mathbf{a}^{[L]} + \mathbf{b}^{[L+1]})
$$ 
for the last layer, where \( g_{\text{out}} \) is the activation function for the output layer.

Thus, the neural network can be succinctly represented as
\[
\widehat{\mathbf{y}} = f(\mathbf{x}) = f^{[L+1]} \circ f^{[L]} \circ \cdots \circ f^{[2]}\circ f^{[1]} (\mathbf{x})
\]
where each \( f^{[\ell]} \) represents a transformation in the neural network.

In particular, the first layer is
\[
   \mathbf{a}^{[1]} = f^{[1]}(\mathbf{x}) = g^{[1]}(\mathbf{W}^{[1]} \mathbf{x} + \mathbf{b}^{[1]})
\]
%
%
%
If a set of features are eliminated from the input, say, from ${k+1}$-th to $d$-th,
we remove the corresponding columns of the matrix $\mathbf{W}^{[1]}$:
$$
\widetilde{\mathbf{W}}^{[1]}\mathbf x=
\left[
\begin{matrix}
w_{11}&w_{12}&\cdots&w_{1k}\\
w_{21}&w_{22}&\cdots&w_{2k}\\
\vdots&\vdots&\ddots&\vdots\\
w_{h_11}&w_{h_12}&\cdots&w_{h_1k}\\
\end{matrix}
\right]
\left[
\begin{matrix}
x_1\\ \vdots\\ x_k\\
\end{matrix}
\right].
$$
We correspondingly substitute the first layer by
\[
   \widetilde{\mathbf{a}}^{[1]} = \tilde{f}^{[1]}(\mathbf{x}) = g^{[1]}(\widetilde{\mathbf{W}}^{[1]} \mathbf{x} + \mathbf{b}^{[1]}),
\]
and the rest of the neural network (i.e., with $k<d$ features) is the same:
\[
\widehat{\widetilde{\mathbf{y}}} = \widetilde{f}(\mathbf{x}) = f^{[L+1]} \circ f^{[L]} \circ \cdots \circ f^{[2]}\circ \widetilde{f}^{[1]} (\mathbf{x}).
\]

\section{Experimental Results}\label{experiments}

This section presents several examples of the application of the proposed residual variance equality test.
Unbiased models are compared by applying the residual variance equality test under identical conditions using a paired $t$-test.

\subsection{Testing Scheme}\label{sec:testscheme}

Residuals $R_A$ and $R_B$ are extracted after training models $A$ and $B$ using a 10-fold cross-validation. When comparing the variances of the residuals, the null hypothesis $H_0$ is 
\[H_0: \sigma^2_{R_A} = \sigma^2_{R_B}.\]

Each nested model is fitted to a subset of the original dataset, derived by removing each column in the data matrix corresponding to a feature that is not intended to be considered.

In addition to the mean squared error and variance, the Wasserstein distance \citep{kantorovich1939} (with $p=1$) is also included in the tables as a measure of similarity between two $n$-sampled empirical probability distributions $F$ and $G$, using the order statistics, as $$W_1(F,G) = \frac{1}{n} \sum_{i=1} ^n \lvert F_{(i)}-G_{(i)} \rvert.$$ This last metric is considered to compare the empirical distribution of residuals from a fitted model with Kronecker's delta, which represents the ideal model where all residuals are zero. This approach enables a comprehensive comparison between the distributions of the base model and those of its nested derivatives, as the specified metric captures the overall shape of the distribution.

Compute the residuals with the strategy described in Section \ref{NestedModels}. Independence will be tested using a distance correlation statistical test \citep{dcorr}.
For visualization purposes, kernel density estimation (KDE) is used, a non-parametric technique equally attributed to Rosenblatt \citep{rosenblatt1956} and Parzen \citep{parzen1962}, which is one of the most widely used methods to estimate the underlying probability density function of a data set.

To ensure reproducibility, the random seeds are set to 0 unless otherwise specified. For bootstrap and Monte Carlo simulations, random seeds are assigned sequentially from 0 to the total number of samples or iterations, which, in the case of bootstrap, corresponds to the number of dataset samples.

\subsection{Statistical Test Performance}
A Monte Carlo simulation is performed to evaluate the performance of the residual variance equality test used in this study. For this purpose, 10000 iterations are conducted, each producing two normally distributed samples having the same variance and with 1000 elements each. The Type I error rate is determined to be 0.0524, with a test power of 0.9611.

\subsection{Simulated Data}
All described synthetic datasets are generated with 1000 samples. The first simulated dataset, \textit{SimData1}, is produced by considering eight standard Gaussian variables, pairwise correlated with a correlation equal to 0.5 and defining the target as in \eqref{simY}
\begin{eqnarray}\label{simY}
    Y &=& 3\Psi(2X_1 + 4X_2 + 3X_3 + 3X_4) + 3\cos(2X_1 + 4X_2 - 3X_3 - 3X_4) + \epsilon,
\end{eqnarray}
where $\Psi$ is the logistic function and $\epsilon \sim t_3$, that is, a Student's $t$ with three degrees of freedom, producing a noise with heavy tails and finite variance. Thus, only the first four variables are relevant to the model. 

A second simulated dataset, labelled \textit{SimData2}, is produced using the same process in \eqref{simY} to obtain the output variable with eight input variables that satisfy pairwise independence.

A third synthetic dataset, denoted \textit{SimData3}, is generated using a complete polynomial of degree 2 with two variables. The values of each variable are drawn independently from a $\mathcal{N}(0,1)$ distribution. Three polynomial regression models, with degrees ranging from 1 to 3, are trained on this dataset. The equality of their residuals is evaluated using the residual variance equality test and the F-test to compare their respective $p$-values, which are calculated using 500 Monte Carlo runs.

A final synthetic dataset, denoted \textit{SimData4}, is created, comprising 10 pairwise independent features each drawn from a $\mathcal{N}(0,1)$ distribution. The target variable is generated by a two-layer neural network with an architecture of (10, 8, 8, 1), where weights are initialized using a Glorot normal distribution.

\subsubsection{\textit{SimData1} - Feature Selection}
For this experiment and the next one, four neural networks will be considered: one taking all variables ($X^{[\text{all}]} \equiv X$ as the dataset and NN$_{\text{all}}$ as the fitted model), one taking all operative variables $\{X_1,X_2,X_3,X_4\}$ (giving $X^{[\text{op}]}$ as the new dataset and NN$_{\text{op}}$ as the neural network), one doing the same for all non-operative variables $\{X_5,X_6,X_7,X_8\}$ ($X^{[\text{nop}]}$ and NN$_{\text{nop}}$), and one taking variables $\{X_1, X_2, X_3,$ $X_6,X_7,X_8\}$ ($X^{[\text{3op-3nop}]}$ and NN$_{\text{3op-3nop}}$).

The tuning and training processes for each model employ the Glorot normal distribution for the initialization of weights, batches of 32 samples, the Adam optimizer, and two hidden layers, with the number of neurons per layer varying between 1 and 10. The base models using all variables that provide the best results in terms of MSE are those with (8, 3, 7, 1) as a dense layer structure (where the leftmost number represents the input neurons, the rightmost number denotes the output neuron and the remaining numbers indicate neurons in the hidden layers) for the case of \textit{SimData1}. The same number of epochs, fixed at 200, is used for each neural network training during the process of generating residuals.

\begin{figure}[htbp]
    \centering
    \includegraphics[width=0.75\linewidth]{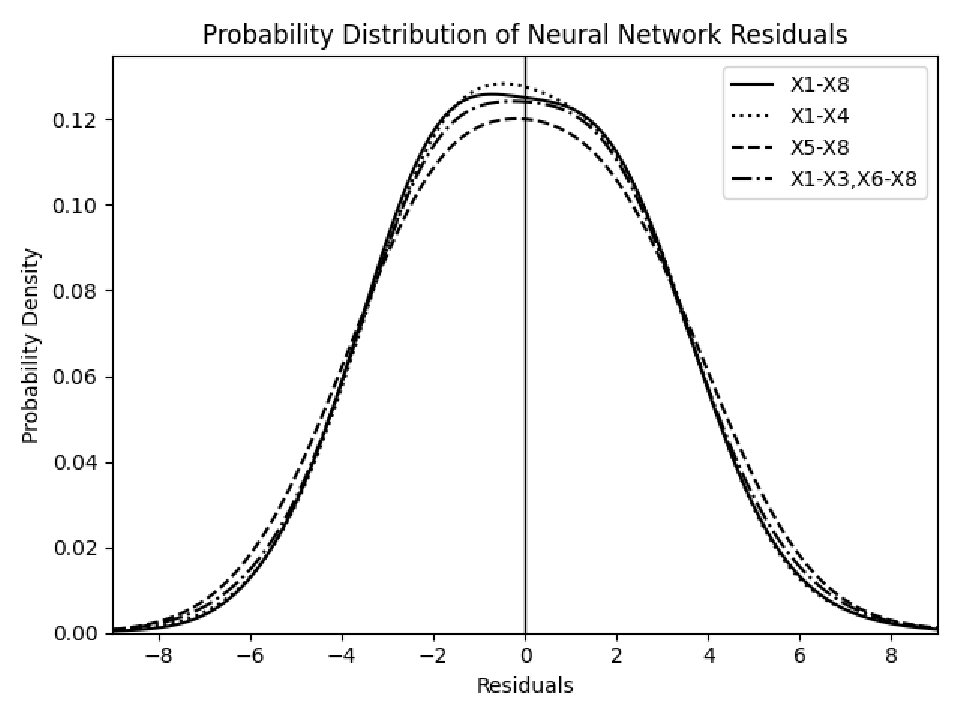}
    \caption{\textit{SimData1}: comparison of the residuals of neural network models for different sets of features.}
    \label{fig:simdata1-figure}
\end{figure}

Tables \ref{tab:simdata1-table} and \ref{tab:simdata1-vareq} represent the results of the statistical and residual variance equality test, respectively, while Figure \ref{fig:simdata1-figure} provides a visual representation of residuals' distributions.

\begin{center}
\begin{table}[htbp]
\begin{tabular}{|c | c c c c c c |}
 \hline
 Model & MSE & $W_1(R, \delta_{ij})$ & Variance & $W_1(R,R_{\text{base}})$ & Bias Test & Indep. Test\\ [0.5ex] 
 \hline
 \hline
 NN$_{\text{all}}$ & 6.447 & 2.113 & 6.446 & 0 & 0.786 & 0.389 \\
 \hline
 NN$_{\text{op}}$ & 6.461 & 2.092 & 6.461 & 0.059 & 0.847 & 0.6 \\
 \hline
 NN$_{\text{nop}}$ & 7.354 & 2.243 & 7.354 & 0.154 & 0.818 & 0.636 \\
 \hline
 NN$_{\text{3op-3nop}}$ & 6.837 & 2.159 & 6.837 & 0.072 & 0.837 & 0.271 \\
 \hline
\end{tabular}
\caption{\textit{SimData1}: neural network model evaluation and residual distribution measures. NN$_{\text{all}}$ (all 8 features), NN$_{\text{op}}$ (operative features), NN$_{\text{nop}}$ (non-operative features), NN$_{\text{3op-3nop}}$ (3 first operative features along with the 3 last non-operative features). $W_1$ stands for the Wasserstein distance with $p=1$. The last two columns are the $p$-values for the bias and independence tests.}
\label{tab:simdata1-table}
\end{table}
\end{center}

\begin{table}[htbp]
\centering
\begin{tabular}{c|c}
                                   
  NN$_{\text{op}}$ & 8.834$\times 10^{-1}$                    \\ 
  NN$_{\text{nop}}$ & 5.323$\times 10^{-7}$ \\ 
  NN$_{\text{3op-3nop}}$ & 2.082$\times 10^{-3}$ \\ 
  \hline
  $\sigma^2_A = \sigma^2_B$ & NN$_{\text{all}}$ \\ 
\end{tabular}
\caption{\textit{SimData1}: neural network residual variance equality test $p$-values.}
\label{tab:simdata1-vareq}
\end{table}

\subsubsection{\textit{SimData2} - Feature Selection}
\begin{figure}[htbp]
    \centering
    \includegraphics[width=0.75\linewidth]{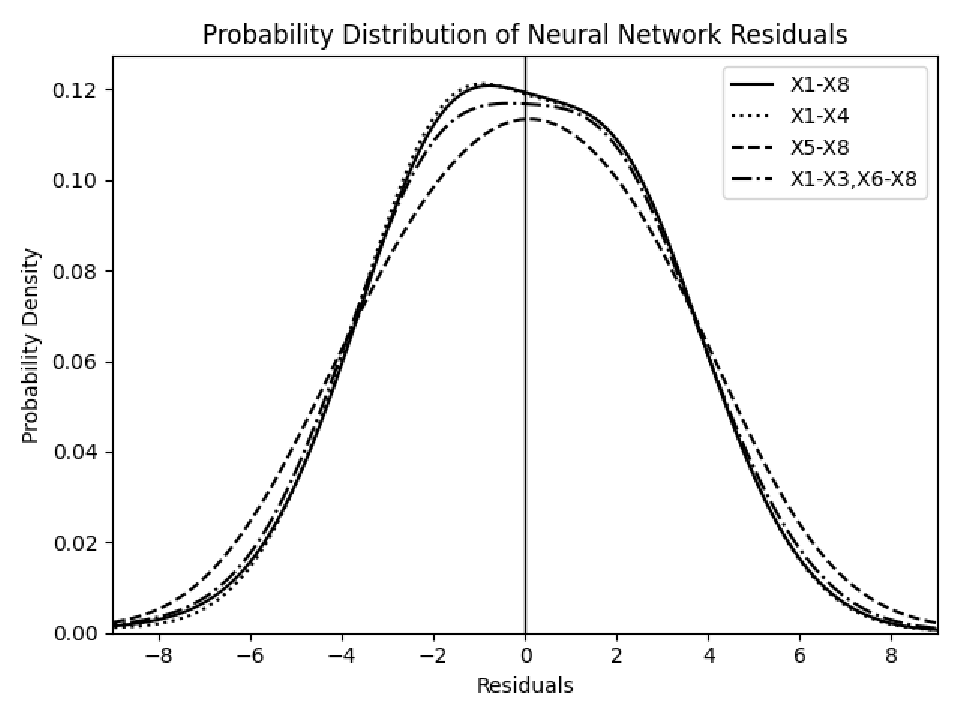}
    \caption{\textit{SimData2}: comparison of the residuals of neural network models for different sets of features.}
    \label{fig:simdata2-figure}
\end{figure}

The profiled base model yielding the most favourable results in terms of MSE is characterised by a dense layer architecture of (8, 2, 6, 1). Model fitting is performed in the same manner as in the preceding experiment, with the number of training epochs fixed at 200. The statistical results of this experiment are collected in Tables \ref{tab:simdata2-table} and \ref{tab:simdata2-vareq}, as well as illustrated in Figure \ref{fig:simdata2-figure}.

\begin{center}
\begin{table}[htbp]
\begin{tabular}{|c | c c c c c c |}
 \hline
 Model & MSE & $W_1(R, \delta_{ij})$ & Variance & $W_1(R,R_{\text{base}})$ & Bias Test & Indep. Test\\ [0.5ex] 
 \hline
 \hline
 NN$_{\text{all}}$ & 7.587 & 2.25 & 7.583 & 0 & 0.952 & 0.454 \\
 \hline
 NN$_{\text{op}}$ & 7.541 & 2.249 & 7.536 & 0.055 & 0.984 & 0.303 \\
 \hline
 NN$_{\text{nop}}$ & 9.096 & 2.437 & 9.089 & 0.232 & 0.973 & 0.513 \\
 \hline
 NN$_{\text{3op-3nop}}$ & 8.173 & 2.326 & 8.166 & 0.129 & 0.74 & 0.871 \\
 \hline
\end{tabular}
\caption{\textit{SimData2}: neural network model evaluation and residual distribution measures. NN$_{\text{all}}$ (all 8 features), NN$_{\text{op}}$ (operative features), NN$_{\text{nop}}$ (non-operative features), NN$_{\text{3op-3nop}}$ (3 first operative features along with the 3 last non-operative features). $W_1$ stands for the Wasserstein distance with $p=1$. The last two columns are the $p$-values for the bias and independence tests.}
\label{tab:simdata2-table}
\end{table}
\end{center}

\begin{table}[htbp]
\centering
\begin{tabular}{c|ccc}
                                   
  NN$_{\text{op}}$ & 8.792$\times 10^{-2}$                    \\ 
  NN$_{\text{nop}}$ & 1.970$\times 10^{-9}$ \\ 
  NN$_{\text{3op-3nop}}$ & 4.571$\times 10^{-2}$ \\ 
  \hline
  $\sigma^2_A = \sigma^2_B$ & NN$_{\text{all}}$ 
\end{tabular}
\caption{\textit{SimData2}: neural network residual variance equality test $p$-values.}
\label{tab:simdata2-vareq}
\end{table}

\subsubsection{\textit{SimData3} - Model Selection}

This study aims to assess the results of the residual variance equality test in comparison with those obtained from the F-test applied to polynomial regression. To this end, a dataset of 1000 samples comprising two independent features, each generated from a $\mathcal{N}(0,1)$ distribution, is constructed using a complete second-degree polynomial to calculate the target variable \eqref{poly2}.
\begin{eqnarray}\label{poly2}
    Y = 1 + X_0 + X_1 + X_0^2 + X_0X_1 + X_1^2 + \epsilon,
\end{eqnarray}
where $\epsilon \sim t_3$.

Tables \ref{tab:simdata3-table} and \ref{tab:simdata3-vareq-ftest} collect the statistical results on this experiment where there is no complete base model as in the feature selection examples above.

\begin{center}
\begin{table}[htbp]
\begin{tabular}{|c | c c c c c |}
 \hline
 Model & MSE & $W_1(R, \delta_{ij})$ & Variance & Bias Test & Independence Test\\ [0.5ex] 
 \hline
 \hline
 Poly (deg=1) & 6.646 & 1.936 & 6.646 & 0.994 & 0.822 \\
 \hline
 Poly (deg=2) & 2.583 & 1.108 & 2.583 & 0.878 & 0.328 \\
 \hline
 Poly (deg=3) & 2.595 & 1.112 & 2.595 & 0.886 & 0.389 \\
 \hline
\end{tabular}
\caption{\textit{SimData3}: polynomial regression model evaluation and residual distribution measures. $W_1$ stands for the Wasserstein distance with $p=1$. The last two columns are the $p$-values for the bias and independence tests.}
\label{tab:simdata3-table}
\end{table}
\end{center}

\begin{table}[htbp]
\centering
\begin{tabular}{c|cccc}                             
  Poly (deg=2) & 1.552$\times 10^{-13}$  & & 2.035$\times 10^{-4}$                 \\ 
  Poly (deg=3) & 1.910$\times 10^{-13}$ & 2.359$\times 10^{-1}$ & 2.185$\times 10^{-4}$ & 4.350$\times 10^{-1}$                   \\ \hline
  $\sigma^2_A = \sigma^2_B$ & Poly (deg=1) & Poly (deg=2) & Poly (deg=1) & Poly (deg=2) \\ 
\end{tabular}
\caption{\textit{SimData3}: polynomial regression residual variance equality test mean $p$-values (left) and polynomial regression F-test mean $p$-values (right).}
\label{tab:simdata3-vareq-ftest}
\end{table}

\subsubsection{\textit{SimData4} - Model Selection}

The primary objective of this investigation is to identify instances of underfitting, well-fitting, and overfitting among fitted models of the same family. This analysis is performed using the residual variance equality test and the bias test performed in each execution of the experiments described.

To introduce variability to the target generated by the 2-layered neural network, random noise following a normal distribution, $\mathcal{N}(0,0.05s_y)$, is added to it, where $s_y$ represents the sample standard deviation of the output of the neural network.

\begin{figure}[htbp]
    \centering
    \includegraphics[width=0.75\linewidth]{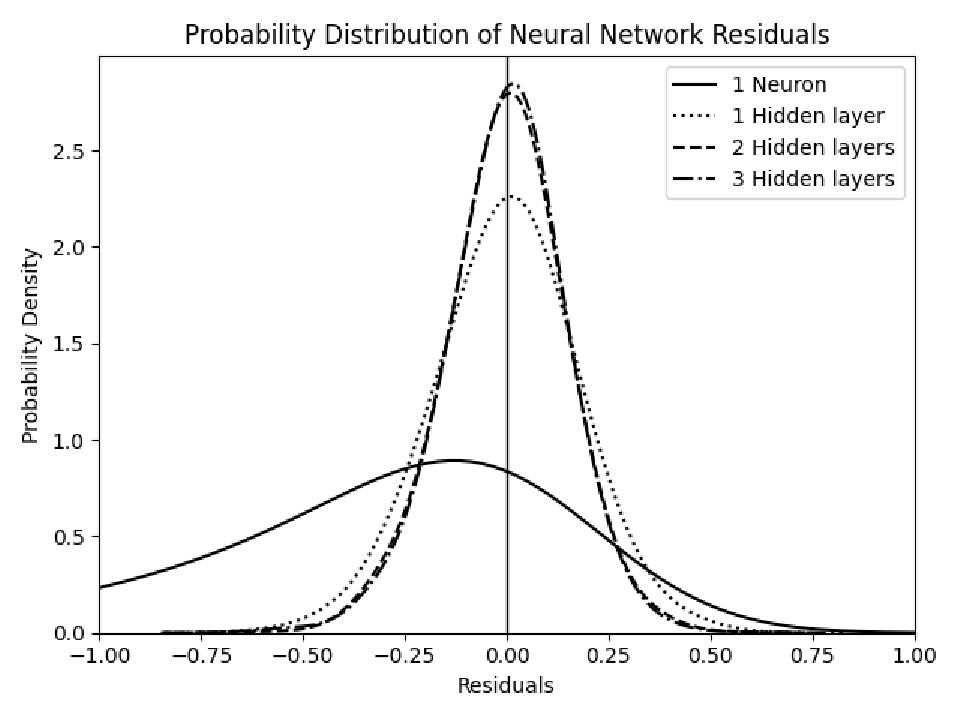}
    \caption{\textit{SimData4}: comparison of the residuals of neural network models for different number of hidden layers.}
    \label{fig:simdata4-figure}
\end{figure}

Four distinct neural networks are initialized using a Glorot normal distribution, with the random seed set to 1, differing from the seed used in the neural network generating the data. These models vary in complexity, with architectures defined as follows: a single artificial neuron, and three neural networks with (10, 8, 1), (10, 8, 8, 1), and (10, 8, 8, 8, 1) as architectures (NN$_{\text{1-layer}}$, NN$_{\text{2-layers}}$, NN$_{\text{3-layers}}$ respectively). Each model is trained using the Adam optimizer, configured with a learning rate of 0.001, batch sizes of 32 samples, and trained over 200 epochs.
Tables \ref{tab:simdata4-table} and \ref{tab:simdata4-vareq}, as well as Figure \ref{fig:simdata4-figure} show the results for this experiment.

\begin{center}
\begin{table}[htbp]
\begin{tabular}{|c | c c c c c |}
 \hline
 Model & MSE & $W_1(R, \delta_{ij})$ & Variance & Bias Test & Independence Test\\ [0.5ex] 
 \hline
 \hline
 1 Neuron & 0.444 & 0.437 & 0.292 & 2.698$10^{-129}$ & 0.263 \\
 \hline
 NN$_{\text{1-layer}}$ & 0.028 & 0.131 & 0.028 & 0.659 & 0.727 \\
 \hline
 NN$_{\text{2-layers}}$ & 0.019 & 0.107 & 0.019 & 0.521 & 0.887 \\
 \hline
 NN$_{\text{3-layers}}$ & 0.019 & 0.105 & 0.019 & 0.550 & 0.099 \\
 \hline
\end{tabular}
\caption{\textit{SimData4}: neural network model evaluation and residual distribution measures. $W_1$ stands for the Wasserstein distance with $p=1$. The last two columns are the $p$-values for the bias and independence tests.}
\label{tab:simdata4-table}
\end{table}
\end{center}

\begin{table}[htbp]
\centering
\begin{tabular}{c|cc}
  NN$_{\text{2-layers}}$ & 8.882$\times 10^{-16}$                    \\ 
  NN$_{\text{3-layers}}$ & 1.332$\times 10^{-14}$ & 8.671$\times 10^{-1}$                 \\
  \hline
  $\sigma^2_A = \sigma^2_B$ & NN$_{\text{1-layer}}$ & NN$_{\text{2-layers}}$ \\ 
\end{tabular}
\caption{\textit{SimData4}: neural network residual variance equality test $p$-values.}
\label{tab:simdata4-vareq}
\end{table}

\subsubsection{Discussion on Simulated Data Results}

Taking $\alpha=0.05$, all models on \textit{SimData1}, \textit{SimData2} and \textit{SimData3} have centered residuals by a $t$-test and are therefore considered unbiased. Of all models in \textit{SimData4}, the only one being biased is the one using only one neuron, that, together with its poorer performance, represents an example of underfitted model using fewer resources than necessary. 
In Tables \ref{tab:simdata1-vareq} and \ref{tab:simdata2-vareq}, we see that model NN$_{\text{op}}$, which uses only operative features, is the only nested model that statistically maintains the same residual variance as the model using all available features, while all other models exhibit a larger residual variance. In both cases, the selection criterion effectively identifies the truly operative features. The exclusion of an operative feature has a statistically significant negative impact even using more input variables than NN$_{\text{op}}$, leading to a higher MSE and an increase in the variance of the residuals, which was expected from \eqref{eq_var}. This effect is also observable through the Wasserstein distance, which quantifies the divergence of the nested model from the base model.
Regarding the model selection examples, all models in \textit{SimData3} agree with the residual variance equality test and the F-test, both indicating equivalent variance for the quadratic and cubic models. Consequently, the second-degree model is preferred due to its greater simplicity.
As for \textit{SimData4}, the neural network with a single hidden layer, while not biased, demonstrates inferior performance compared to the two neural networks with additional layers. Specifically, the three-layered neural network outperforms its two-layered counterpart in terms of mean squared error, as indicated by the Wasserstein distance metric. However, despite the three-layered model being favored by accuracy measures, the two-layered model is preferable for better generalization performance and is the option selected by the proposed criterion as both exhibit the same residual variance by the residual variance equality test.

\subsection{Real Data}

\subsubsection*{Boston Housing Dataset}
This dataset contains information collected by the U.S. Census Service on housing in the Boston area \citep{harrison1978}. It has been used extensively throughout the literature to benchmark algorithms. The dataset consists of 506 observations, each with 13 features that accompany a target variable. 
The main task is to predict the median value of owner-occupied homes (variable \verb+MEDV+).
\verb+CHAS+, \verb+RAD+ and \verb+B+ features are removed in the preprocessing stage due to being categorical\footnote{In this study, experimentation will be carried out on continuous data.} and controversial\footnote{The B feature, deriving from a formula using the proportion of black people by town has been subject to ethical debates and removed from Scikit-learn library, where it previously was available.} attributes. \verb+MEDV+ target attribute possibly has a censoring value of $50.0$, which is deduced by comparing the number of samples with this value with respect to the number of them with values between $40.0$ and $50.0$\footnote{Delve Datasets, https://www.cs.toronto.edu/\textasciitilde delve/data/boston/bostonDetail.html, University of Toronto. Last updated: October 10, 1996.}. The target is taken to be a log-transformed \verb+MEDV+ and outliers are selected with Tukey's Interquartile Range method \citep{tukey1977}, with all samples with 2 or more outliers removed.

\subsubsection{Boston Housing Dataset - Feature Selection}
\begin{figure}[htbp]
    \centering
    \includegraphics[width=0.75\linewidth]{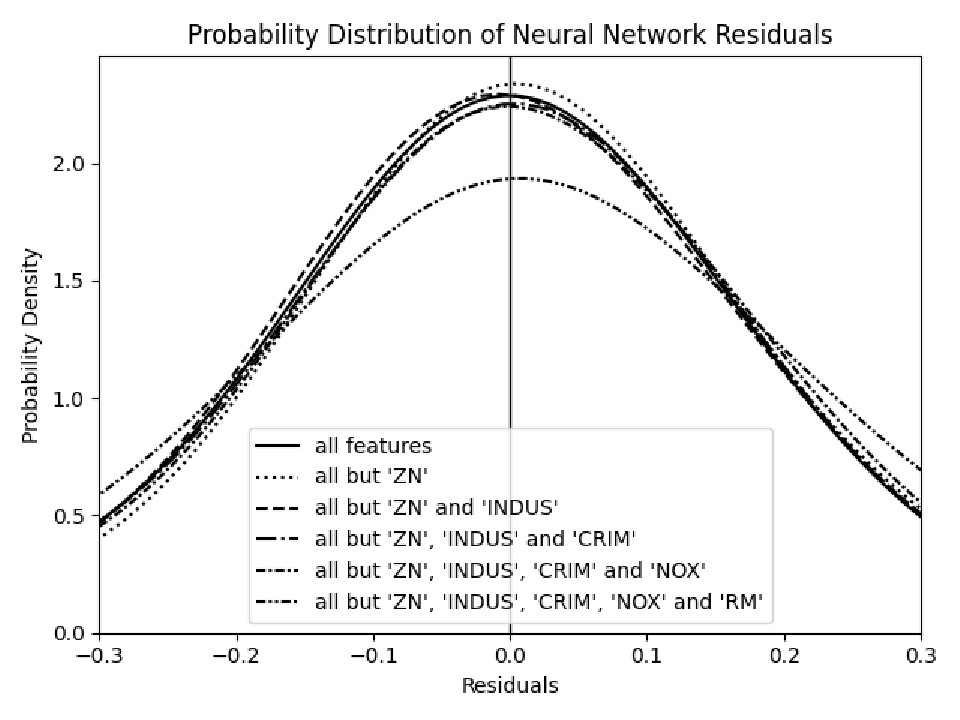}
    \caption{Boston Housing Dataset: comparison of residuals of neural network models for different sets of features.}
    \label{fig:boston-nn-figure}
\end{figure}

An initial experiment is conducted to evaluate the performance of neural networks using a preprocessed Boston Housing dataset that incorporates 10 input features. The neural network architectures are configured after a previous tuning with a structure of (10, 5, 13, 1), where the inner numbers represent the units in the hidden layers. The weights of the neural network models are initialized using a Glorot normal distribution. The models are trained over 200 epochs employing the Adam optimizer with batch sizes of 8 samples. Relevant statistical results from this investigation are presented in Tables \ref{tab:boston-nn-table} and \ref{tab:boston-nn-vareq}, while Figure \ref{fig:boston-nn-figure} provides a graphical representation of the approximate distributions of the residuals.

\begin{center}
\begin{table}[htbp]
\begin{tabular}{|c | c c c c c c |}
 \hline
 Model & MSE & $W_1(R, \delta_{ij})$ & Variance & $W_1(R,R_{\text{base}})$ & Bias Test & Indep. Test\\ [0.5ex]
 \hline
 \hline
 NN$_{\text{all}}$ & 0.031 & 0.125 & 0.031 & 0.0 & 0.654 & 0.431\\
 \hline
 NN$_{\text{-1feature}}$ & 0.03 & 0.123 & 0.03 & 0.011 & 0.616 & 0.651\\
 \hline
 NN$_{\text{-2features}}$ & 0.032 & 0.124 & 0.032 & 0.008 & 0.768 & 0.553\\
 \hline
 NN$_{\text{-3features}}$ & 0.035 & 0.127 & 0.035 & 0.007 & 0.623 & 0.264\\
 \hline
 NN$_{\text{-4features}}$ & 0.031 & 0.128 & 0.031 & 0.009 & 0.973 & 0.837\\
 \hline
 NN$_{\text{-5features}}$ & 0.038 & 0.146 & 0.038 & 0.024 & 0.587 & 0.527\\
 \hline
\end{tabular}
\caption{Boston Housing Dataset: neural network model evaluation and residual distribution measures. NN$_{\text{all}}$ (all features), NN$_{\text{-1feature}}$ (all but ZN feature), NN$_{\text{-2features}}$ (all but ZN and INDUS features), NN$_{\text{-3features}}$ (all but ZN, INDUS and CRIM features), NN$_{\text{-4features}}$ (all but ZN, INDUS, CRIM and NOX features) and NN$_{\text{-5features}}$ (all but ZN, INDUS, CRIM, NOX and RM features). $W_1$ stands for the Wasserstein distance with $p=1$. The last two columns are the $p$-values for the bias and independence tests.}
\label{tab:boston-nn-table}
\end{table}
\end{center}

\begin{table}[htbp]
\centering
\begin{tabular}{c|ccccc}
                              
  NN$_{\text{-1feature}}$ & 6.152$\times 10^{-1}$ \\ 
  NN$_{\text{-2features}}$ & 8.337$\times 10^{-1}$ & 5.648$\times 10^{-1}$ \\
  NN$_{\text{-3features}}$ & 3.114$\times 10^{-1}$ & 1.886$\times 10^{-1}$ & 2.533$\times 10^{-1}$ \\
  NN$_{\text{-4features}}$ & 9.641$\times 10^{-1}$ & 6.635$\times 10^{-1}$ & 8.247$\times 10^{-1}$ & 2.844$\times 10^{-1}$ \\
  NN$_{\text{-5features}}$ & 1.947$\times 10^{-2}$ & 4.099$\times 10^{-3}$ & 1.823$\times 10^{-1}$ & 5.541$\times 10^{-1}$ & 7.770$\times 10^{-4}$ \\
  \hline
  $\sigma^2_A = \sigma^2_B$ & NN$_{\text{all}}$ & NN$_{\text{-1feature}}$ & NN$_{\text{-2features}}$ & NN$_{\text{-3features}}$ & NN$_{\text{-4features}}$ \\ 
\end{tabular}
\caption{Boston Housing Dataset: neural network residual variance equality test.}
\label{tab:boston-nn-vareq}
\end{table}

\subsubsection{Discussion on Feature Selection with Boston Housing Dataset}

At a significance level of $\alpha=0.05$, the neural network models fitted in these experiments are unbiased and have statistically independent residuals. The proposed approach contrasts with the outcome of the commonly applied mean squared error criterion, which would have resulted in the exclusion of only the \verb+ZN+ feature, as all other models exhibit a higher mean squared error than the corresponding model, although the model with the best generalization capacity among those studied is the one that does not take into account the first four features.

\section{Conclusions}\label{sec:conclusion}
The residual variance equality test serves as a robust alternative to traditional model selection methods, such as the minimum MSE criterion, by prioritizing information-dense relationships between input data and target variables. This leads to models that are more reliable and resistant to noise and irrelevant data features compared to those selected by conventional machine learning heuristics.  
Unlike performance-based selection, this method statistically emphasizes feature importance, assessing each input feature. Some uses of this test include algorithmic searches for maximally compact models derived from an initial parent model.

\section{Disclosure statement}\label{disclosure-statement}

The authors report that there are no competing interests to declare. 

\section{Data Availability Statement}\label{data-availability-statement}

All data used in this study are publicly available.

\end{document}